\documentclass[journal]{IEEEtran}

\usepackage{times}
\usepackage{epsfig}
\usepackage{graphicx,times}
\usepackage{subfigure}  
\usepackage{amsmath}
\usepackage{amssymb}
\usepackage{booktabs}
\usepackage{epstopdf}
\usepackage[dvipsnames, svgnames, x11names]{xcolor}
\usepackage{multirow}

\newcommand{\squishlist}{
	\begin{list}{$\bullet$}
		{ \setlength{\itemsep}{0pt}
			\setlength{\parsep}{3pt}
			\setlength{\topsep}{3pt}
			\setlength{\partopsep}{0pt}
			\setlength{\leftmargin}{1.5em}
			\setlength{\labelwidth}{1em}
			\setlength{\labelsep}{0.5em}}
	}
	
	\newcommand{\squishend}{
\end{list}  }

\ifCLASSINFOpdf
\else
\fi
\begin{document}
%
\title{ALR-GAN: Adaptive Layout  Refinement for Text-to-Image Synthesis}

\author{Hongchen Tan,    Baocai Yin*, Kun Wei,  Xiuping Liu,  and  Xin Li

	\thanks{(Corresponding Author: Baocai Yin)
		
		Hongchen Tan and Baocai Yin are  with Beijing Institute of Artificial Intelligence,  Beijing University of Technology, Beijing 100124, China (e-mail: tanhongchenphd@bjut.edu.cn;  ybc@bjut.edu.cn.).
		
        Xin Li is with Section of Visual Computing and Creative Media, School of Performance, Visualization, and Fine Arts, Texas A\&M University,  College Station, Texas 77843,  United States of America (e-mail: xinli@tamu.edu).
        
        Xiuping Liu is with School of Mathematical Sciences, Dalian University of Technology, Dalian 116024, China (xpliu@dlut.edu.cn.). 
               
		Kun Wei is the School of Electronic Engineering, Xidian University, Xi’an 710071, China (e-mail: weikunsk@gmail.com).

	}
	\thanks{}
	\thanks{}}

%

\maketitle
\begin{abstract}
We propose a novel  Text-to-Image Generation Network,  Adaptive Layout  Refinement Generative Adversarial Network (ALR-GAN), to adaptively refine  the  layout of synthesized images  without   any auxiliary information. 
The  ALR-GAN includes an  Adaptive Layout  Refinement (ALR) module   and  a   Layout Visual Refinement (LVR) loss.
The ALR module    aligns  the layout structure (which refers to  locations of   objects and background) of a synthesized image with  that  of   its corresponding real  image. 
In ALR module, we  proposed  an  Adaptive Layout  Refinement (ALR) loss  to balance the matching of hard   and easy features, for more efficient layout structure matching. 
Based on the refined layout structure,  the LVR loss     further refines the visual representation within the  layout area.
Experimental results on two widely-used datasets show that   ALR-GAN  performs competitively  at the Text-to-Image generation task.
\end{abstract}


\section{Introduction}\label{Introduction0}

Text-to-Image Generation (T2I) aims to  synthesize  photorealistic images from a   text description.    
To realize this challenging cross-modal  generation  task   can facilitate   multimedia tasks   such as image  editing~\cite{LisaiACMMM2020, yahuiACMMM2020},  story  visualization~\cite{StoryGAN2019}, and cross-modal retrieval~\cite{Jiuxiang_cvpr2018}.
Owing to      Generative Adversarial Networks (GANs)~\cite{Goodfellow2014Generative}, the latest T2I methods  facilitate the  synthesis of  high-resolution images~\cite{Han2017StackGAN, Han2018StackGANn, Zhang2018Photographic}, refinement of image details~\cite{Xu2017AttnGAN,   Minfeng2019,  Bowen2020, 8989803}, and enhancement of image semantics~\cite{thc2019, 2019Tingting, Guojun2019, 9552559, 8890866}. 

\begin{figure}[h!tb]
	\centering
	\includegraphics[scale=0.3]{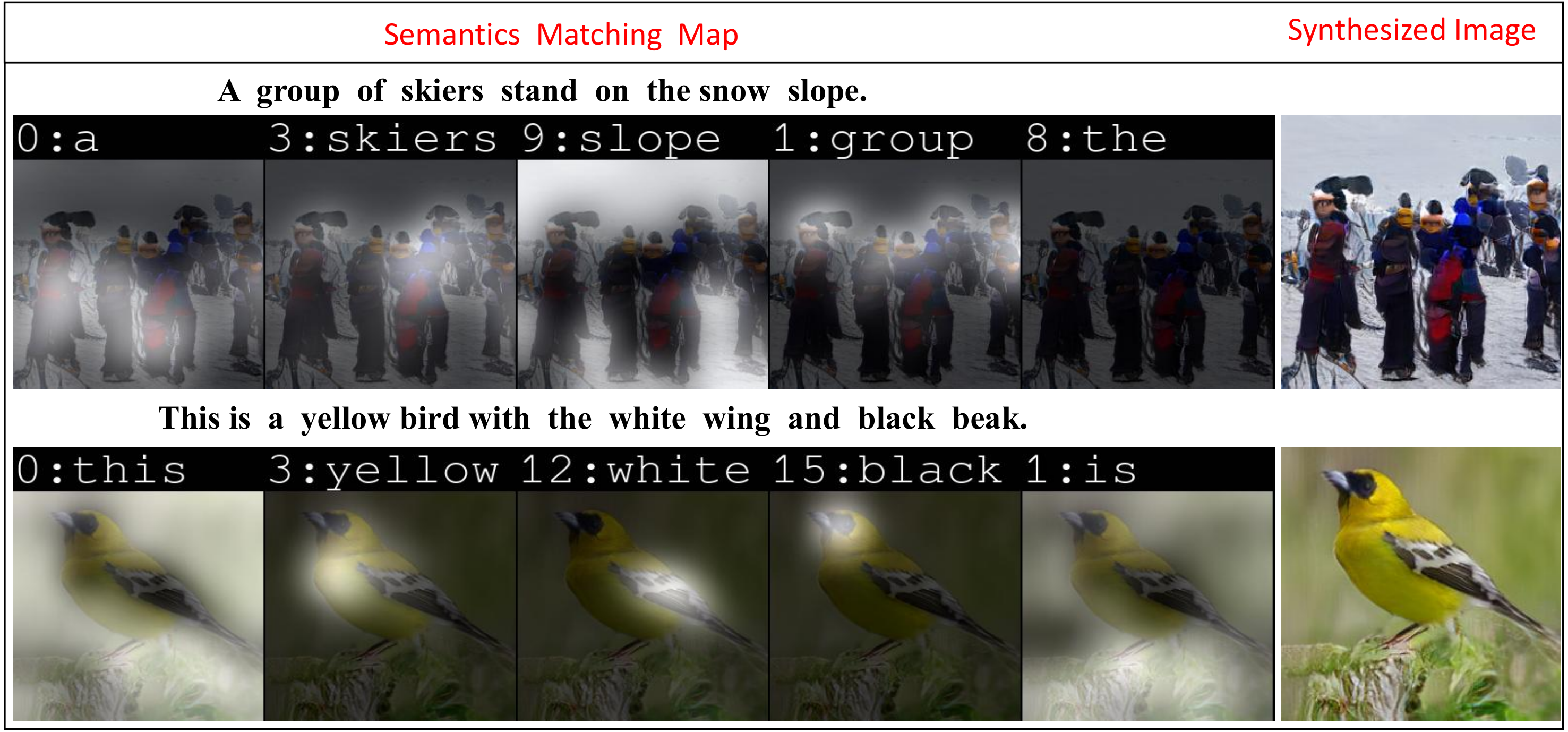}
	\caption{Visualization  of semantics-matching  maps  and    corresponding synthesized  image.  The brighter the color, the more semantic correlation between the  word and image subregions. } 
	\label{Attention_Map} 
\end{figure}

While these   T2I methods  can synthesize high-quality images,  they tend to  focus  on  single-object synthesis, such as a bird, flower, or dog. 
For    complex image synthesis tasks such as those on the MS-COCO dataset~\cite{Lin2014Microsoft},   synthesized  objects are easily  placed on  various  unreasonable locations of the image, i.e.,  the layout  structure (which refers to the locations of the objects and background) is chaotic. 
Some excellent methods~\cite{Justin2018Image, Liang2020CPGANCG,  Semantic_c_2020, Wenb19cvpr} have improved  the layout structure through  auxiliary information such as the object bounding box,   object shape, and scene graph. However, 1) the acquisition of this auxiliary information is generally expensive and  
not conducive to the promotion and application of the task; and 
2)    these methods generally ignore the visual quality  within the layout area. 
We aim to  improve   the layout of  the synthesized image with no  auxiliary information.

We    exploit the layout structure information from  the  image and    text description  to  improve  the  layout structure  of the  synthesized image.
In  the Text-to-Image  matching task~\cite{Faghri2017VSE, Gu2017Look, Zheng2018Dual},   each  word  has   semantics-matched  subregions on its  corresponding image  by      matching semantics between  words and    image  subregions.  
In  Fig.~\ref{Attention_Map},  highlighted areas indicate the semantic correlation between the  word and corresponding image  subregions, reflecting the semantic location  or structure information of the word on the image. 
Accordingly, all semantics-matched  subregions of   words  in the text description  reflect the  layout structure of the corresponding image. 
Therefore, the obvious idea is to align  the layout structure of the synthesized  image with  that  of  its  corresponding  real image based on semantics-matching. 
However, in the  layout structure matching process,  the structures of some  subregions are easier to align than   others. 
Such hard  subregions cause major difficulty in layout  refinement. 
Thus, in each training stage, the model should spend more effort in layout structure  matching in such hard subregions. 
For this, we design an adaptive weight adjustment mechanism   to  adaptively improve  the hard   and    easy structures  for  the synthesized image.

In addition to improving the layout structure, we      enhance  the  visual representation of   the  synthesized image.
One of the most straightforward ideas is to directly constrain the visual  consistency between  the  whole synthesized  image and the whole real image.
However, the semantics of the  text description  only covers some  partial  semantics of the images in  general. 
Over-constraint on layout and details that are not included in the text description can significantly increase the training burden of the model. 
Thus,  within the corrected  layout area,  we try  to  align  the   visual  representations   of the synthesized  and corresponding real images.

Finally,  we propose  an    Adaptive Layout  Refinement Generative Adversarial Network (ALR-GAN) to improve the layout of  the synthesized image.
The ALR-GAN  includes an  Adaptive Layout  Refinement (ALR) module   and     Layout Visual Refinement (LVR) loss.
The  ALR module and  the proposed Adaptive Layout  Refinement (ALR) loss act to adaptively  align the layout structure  of the  synthesized image  with   the   visual  representation  of  its  corresponding  real image.  
The   adaptive weight adjustment mechanism in the  ALR  loss  adaptively balances the matching  weights of hard and easy parts in the layout  structure alignment process.
In  the layout  refined by the ALR module, the  LVR loss   is  designed   to  align the  texture perception and style information of the  synthesized image  with  that  of its corresponding real image. 
Our contributions can be summarized as follows:

$\bullet$ We propose an ALR module,  equipped with the proposed  ALR loss  to  adaptively  refine   the layout structure  of synthesized images.

$\bullet$ We propose    LVR loss  to  enhance  the  visual representation within the layout.

$\bullet$ Experimental results and analysis show the efficacy of ALR-GAN on the  CUB-Bird \cite{WahCUB_200_2011} and large-scale MS-COCO \cite{Lin2014Microsoft} benchmarks based on four metrics.

\section{Related Work}\label{RelatedWork}

In current T2I models, the design and refinement of layouts of synthesized objects and details are often achieved with the help of auxiliary information. 
\emph{Visual Question Answering (VQA)} datasets are used to enhance object semantics during image synthesis~\cite{Leveraging_2020, Chatpainter_2018, Niu_VAQ_2020, Tobias2019}.  
These models   extract semantic information from   questions and    feed it to the VQA model. 
Visual question-answering loss  is adopted to align the semantics in generated images with those extracted in the VQA tasks. 
Another type of auxiliary information, the object box  \cite{Sylvain_2020, Wu_Sun_2019, Leveraging_Sun_2019, Leveraging_layout_2019}, helps control the structure of objects in   synthesized scenes.  The
object box defines the location of each object using a bounding box with a category label. 
It can be used to control the placement of these objects during image synthesis from a higher-level perspective. 
Object shape information is another type of auxiliary information that has been used~\cite{Hong2018Inferring, Wenb19cvpr} to help generator networks improve the geometric properties of synthesized objects.
In text description, the relationship between multiple objects can often be more explicitly represented using a more structured scene graph. 
Compared with the plain text description, scene graphs can encode more information on objects' positions and spatial relationships, which can provide   explicit guidance in layout design and refinement~~\cite{Justin2018Image, Visual_relation_2020,  Advances_Convolutional_2019}.

While   auxiliary information can effectively help the layout design, it must be obtained through extra data collection and annotation procedures (which are often expensive and not scalable) or by solving additional tasks (which might not be easier than T2I itself). 
Unlike the above, we  exploit layout  structure information from the  text description and corresponding images. Layout design and refinement can then be performed together with the mainstream T2I pipelines, without  collecting extra auxiliary information.

\section{Adaptive Layout  Refinement Generative Adversarial Network }

\begin{figure}[h!tb]
	\centering
	\includegraphics[scale=0.44]{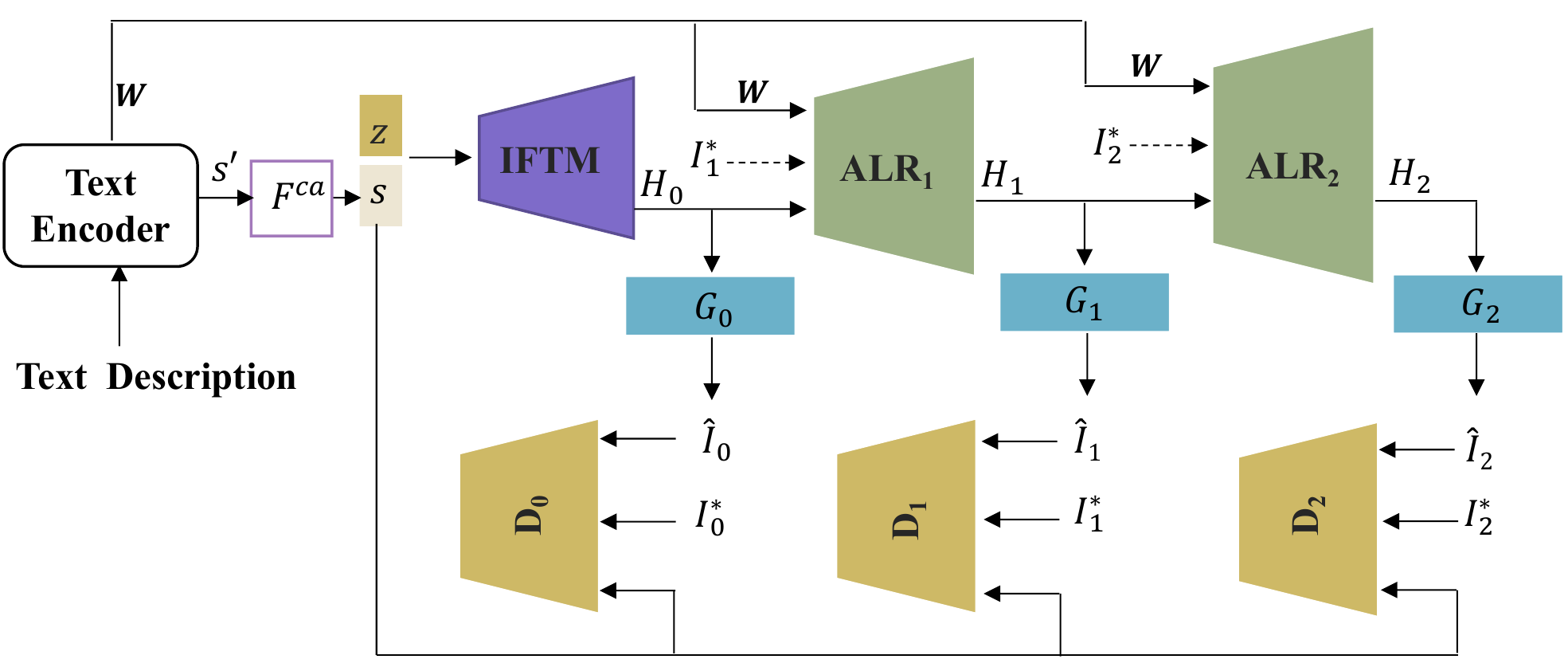}
	\caption{Architecture of   proposed ALR-GAN.   } 
	\label{Fig1:Pipeline} 
\end{figure}

Fig.~\ref{Fig1:Pipeline} shows the architecture of the proposed ALR-GAN, which  contains two new components: the  Adaptive Layout  Refinement (ALR) module   and     Layout Visual Refinement (LVR) loss.   
The ALR module is equipped with the proposed   Adaptive Layout  Refinement (ALR)  loss to  adaptively  refine  the  layout  structure  of  a synthesized image with the  aid  of  that of  its corresponding real  image. 
LVR  loss  aims  to  enhance  the   texture perception and style information  within  the layout area.

We provide an overview of the ALR-GAN structure in  section~\ref{Overview}, describe the  ALR module  in  section~\ref{ALR},   describe  the LVR loss in  section~\ref{LVR_loss}, and  summarize  all the  losses  for ALR-GAN in section~\ref{Object_fucntions}.

\subsection{Overview}~\label{Overview} 
As shown in Fig.~\ref{Fig1:Pipeline},  ALR-GAN  contains  a  text encoder~\cite{Xu2017AttnGAN}; a conditioning augmentation module~\cite{Han2017StackGAN} $F^{ca}$; $m$ generators $G_{i}$, $\emph{i} =0, 1, 2, \ldots, m-1$);  an  Initial Feature Transition  Module (IFTM); 
$m-1$ ALR modules  $ALR_{i}$, $\emph{i} =1, 2, \ldots, m-1$;  and $m$ discriminators  $D_{i}$, $\emph{i} =0, 1, 2, \ldots, m-1$.

The text encoder transforms the input text description (a single sentence) into  a sentence feature $s'$ and word features $W$.
$F^{ca}$~\cite{Han2017StackGAN} converts   $s'$ to a conditioning sentence feature $s$.  
The IFTM  translates   the text  embedding   $s$  and   noise $z \sim N(0, 1)$ into the  image  feature $H_{0}$,  which   is   an initial generation stage. 
The ALR  module adaptively refines the layout  structure  of   synthesized   images  \emph{in the  training process}.
In  the training stage,  the  input   information  of   $ALR_{i}$ consists of   the  word  features  $W$,     image  feature  $H_{i-1}$, and     $i^{th}$ scale  real  image $I_{i}^{*}$.  The  ouput information  of  $ALR_{i}$  is    the image feature $H_{i}$. 
In the testing stage,   $ALR_{i}$ takes  the image  feature $H_{i-1}$  and    word  feature $W$ to produce  the  hidden feature $H_{i}$. Note that during the testing stage, the input information to the generator  does not include real images.

\subsection{Adaptive Layout  Refinement  module}~\label{ALR}
\begin{figure}[h!tb]
	\centering
	\includegraphics[scale=0.54]{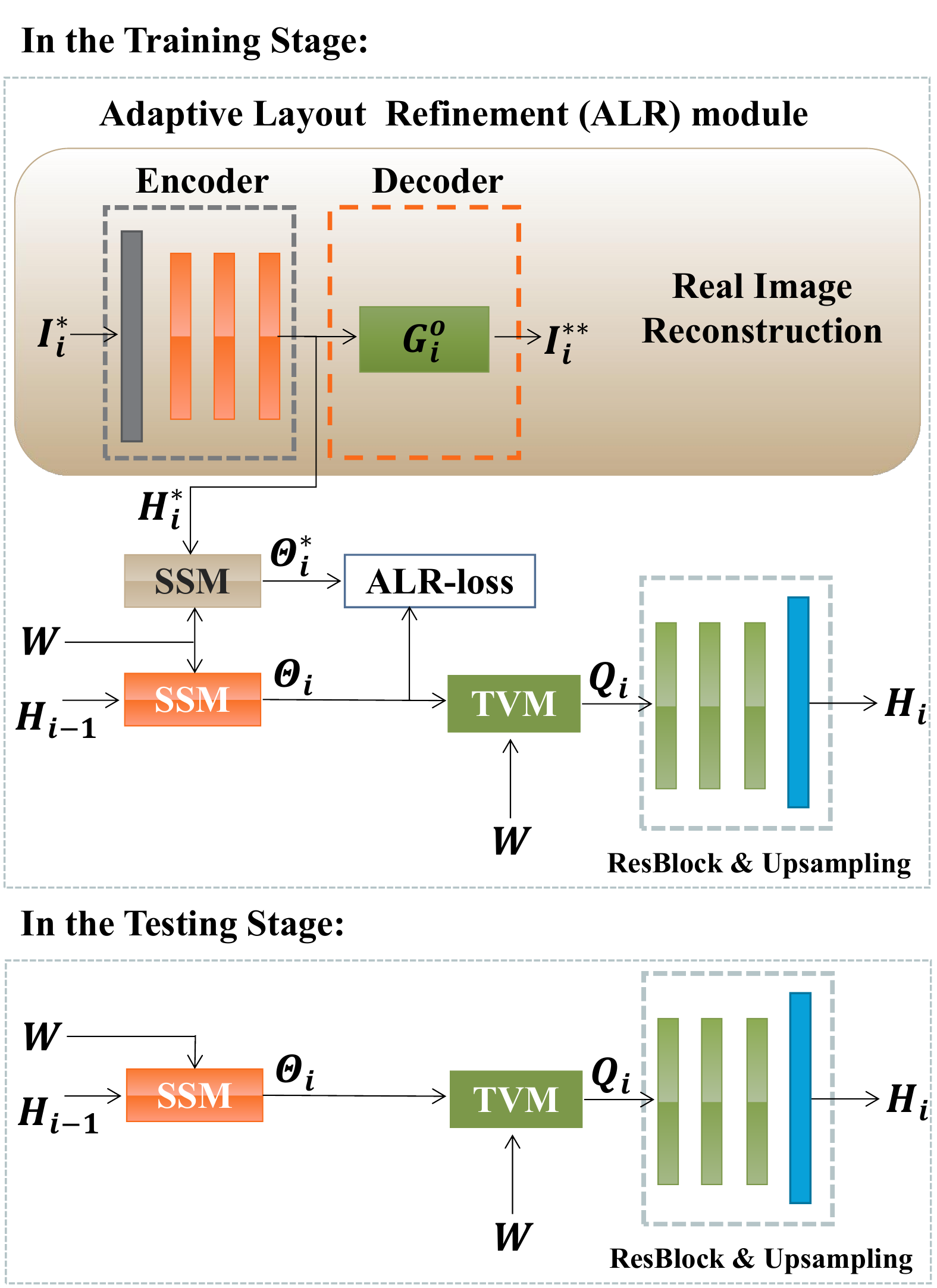}
	\caption{Architecture of ALR module in    training and testing stage.  SSM: Semantics Similarity Matrix;    ALR loss: Adaptive Layout  Refinement loss;   TVM:  Text-Vision  Matrix.  In  the ResBlock \&  Upsampling part,  the  green box is the ResBlock module, and  the   Upsampling module is blue. In the training stage, the ALR module requires the participation of real images, which provide the  high-quality layout information for the  generator; in the testing stage, the input  information  of the ALR module does not include the  real image.} 
	\label{ALR_F} 
\end{figure}

Without the aid  of  auxiliary information, we need  to exploit the  layout structure information from text and images,  and refine the layout structure  of the synthesized image.
As  described in section~\ref{Introduction0}, we can  obtain  the layout structure information of the image from   the semantics-matching between words and   image subregions. 
To this end,   we can  align the layout structure  of the  synthesized image  with  that  of its  corresponding   real image.
To achieve  this goal,  we propose  the Adaptive Layout  Refinement (ALR) module, whose  architecture    is  shown  in Fig.~\ref{ALR_F}.

\subsubsection{Construction of Layout Structure}
In the  $i^{th}$  ALR$_i$  module,   
the word  embeddings  are denoted as  $W=\{w_{j} \in \mathbb{R}^{D} | j=1,2,\cdots,T\}$, the  image feature  of  the previous stage  $ALR_{i-1}$ or IFTM    is denoted as $H_{i-1}=\{h^{k}_{i-1} \in \mathbb{R}^{D} | k=1,2,\cdots,N=W_H \times H_H \}$, $N$ is the number of image subregions ($W_H$ and $H_H$ are the respective width and height of   feature $H_{i-1}$), $D$ is the dimension of the  feature, and  $T$ is the number of words.

We  capture  the layout  structure  from the   semantics  matching   between  the words $W$  and    image  subregions $H_{i-1}$,  defined  as
\begin{equation}\label{SSM_Equation}
	\theta_{k,j}=\frac{exp(S_{k,j}')}{\sum_{k=1}^{T}exp(S_{k,j}')},  \,\,\,\, S'_{k,j}=(h^{k}_{i-1})^{T}w_{j}.
\end{equation}
Here $\Theta=\{ \theta_{k,j}  | k=1, 2, \cdots, N; j=1,2,\cdots,T\}$  is called a  \textbf{ Semantics Similarity Matrix (SSM)}, $\theta_{k,j}$  is the  semantics similaity  weight  between  the ${j}^{th}$ word $w_i$ and the $k^{th}$ image subregion $h^{k}_{i-1}$, and  the above  calculation  process  is  defined   as  a  function,  $\Theta_{i-1}= SSM(W,H_{i-1})$.  \textit{Without ambiguity, we omit the subscripts $i-1$ of $\Theta$, i.e., $\Theta= SSM(W,H_{i-1})$.}

In     $\Theta$,   each  matrix $\Theta_j \in \mathbb{R}^{W_H \times  H_H}$   calculates  the semantics  similarity   between  the ${j}^{th}$ word $w_j$   and  all   image subregions   $H$. 
When the weight $\Theta_j$ is overlaid on the image, as shown in Fig.~\ref{Attention_Map}, the highlighted  visual regions  express  that  their semantics are related to  the word $w_j$.  
Accordingly,     $\Theta$ encodes  the  layout structure of the image.  
However, the layout structure of the synthesized  image is often chaotic due to incorrect or inadequate understanding of text semantics by the generator. 
Thus,  we  hope  that    $\Theta$  of the synthesized  image is  aligned   with that  of   the real image.  
Before alignment,  we need  to  calculate     $\Theta^*$   of  the real  image $I^*_i$.

As  shown in  Fig.~\ref{ALR_F},  we use  the encoder (which  contains a  series of convolution blocks)  to extract     feature  $H^*_i$ of the  real image  $I^*_i$. 
To  ensure  the  quality of   $H^*_i$,  we put     $H^*_i$  in  $G_i(\cdot)$  to  reconstruct   image $I^{**}_i$, and introduce the reconstruction loss, 
\begin{equation}\label{eq8}
	\mathcal{L}^{REC}_{i}=|| I^{*}_i- I^{**}_i ||_1
\end{equation}
Then    $\Theta^*$   of  the real  image $I^*_i$  is  calculated by  $\Theta^*=SSM(W,H^*_i)$.

\subsubsection{Adaptive Layout  Refinement (ALR) loss}
We  can align      $\Theta$   with  $\Theta^*$ by  minimizing $|| \Theta-\Theta*||_F$.
During  optimization,  some elements in    $\Theta$   and  $\Theta^*$   are easy to match, and some are hard.  
A hard region causes major problems in  the  layout  refinement process.  
Thus, during training, the model should pay more attention to    matching in  hard regions. 
Balancing easy versus hard information is important to improve  the  model's  performance~\cite{2018Xiankai, Lin2017Focal, 2018Xingping}. 
We  propose     Adaptive Layout  Refinement (ALR) loss  to address  this  issue. 
There  are four  steps to  its construction.

\textbf{Step 1.} Calculate  the absolute  residual tensor  $R= Abs.( \Theta^{*} -\Theta)$, $R=\{ r_{i,j}  | i=1, 2, \cdots, N; j=1,2,\cdots,T\}$, where ``Abs''    denotes  the absolute operation.

\textbf{Step 2.} Divide   the  elements   in   $R$  into   hard  and easy parts. 
We set  a   threshold  value $\gamma$.  
Elements  $r_{i,j}<\gamma$   are easy to match, and  elements   $r_{i,j}\geq\gamma$   are hard to match.

\textbf{Step 3.} Define  the  adaptive  weight  adjustment  terms  for  the easy      and  hard  parts.   
Recently, ~\cite{2018Xiankai, Lin2017Focal, 2018Xingping,   Wei2019AdversarialFC}  adaptively adjusted  the loss   weights  for hard and easy samples, but   these   adjustments are suitable for the sample  but not   the feature pixel. 
Thus,   we   design  an  adaptive feature-level weight  adaptation  mechanism   to   adjust  loss  weights   for    easy and hard  matching elements in    $\Theta$   and  $\Theta^*$.   
There  are  four steps \textbf{(3.a-3.d)}  to construct   the  adaptive   weight  adaptation  mechanism. 
We show  the  adaptive   weight adjustment mechanism for the easy part, and the adjustment    for the   hard part   is similar.

\textbf{Step 3.a.}  Keep the elements of $R$ less than $\gamma$, set the rest to $0$, and call it a tensor,  $R'_{easy}$.

\textbf{Step 3.b.}     $R'_{easy} \in \mathbb{R}^{N \times  T}$  is  maped  into  the  same  latent  space as    $H^*$   by    padding with zeros, which is called  $R_{easy}$.

\textbf{Step 3.c.}  The  feature  corresponding  to  the easy  part  $R_{easy}$   on   real  image  feature $H^*$   is  calculated  by  $R_{easy} \odot H^*$.

\textbf{Step 3.d.} The loss weight  of     $R_{easy}$   is denoted as  $\alpha$, which  is  learned  from  $\alpha=\phi_{\alpha}(R_{easy} \odot H^*)$. Here, $\phi_{\alpha}(\cdot)$  is composed of   a  series  of  neural  layers  and activation layers, and $\odot$   denotes   the  Hadamar Product.   Note  that    $\alpha$   has  the same dimension as   $R$.

For  the  hard   part, we keep the elements of $R$ larger than $\gamma$, set the rest to $0$, and call it    $R_{hard}$.
The loss weight  of     $R_{hard}$    is denoted as  $\beta$, and this is  learned  from  $\beta=\phi_{\beta}(R_{hard}\odot H^*)$.  
$\phi_{\beta}(\cdot)$  is  also composed of  a  series  of  neural    and activation layers.

\textbf{Step 4.} In   the  training   process,  the hard  part  should be  better  focused on.  Therefore, the weight $\beta$ should  be   bigger than  $\alpha$. 
So,  we  design  a  regularization item $softplus( \max(\alpha)-\min(\beta))$ in $L_{ALR}$ to satisfy it.  Here, $y=softplus(x)=\ln(1+e^x)$  is a monotonically increasing function that helps numerical optimization by avoiding negative loss values.

Based on  \textbf{Steps 1--4},   the  ALR loss is defined  by 

\begin{small}
	\begin{equation}\label{AACL_Eq}
		\begin{aligned}
			L^{ALR}_{i}=\frac{1}{N \cdot D}( || \alpha_{i} \odot & R_{easy_{i}} ||_F +  ||\beta_{i} \odot R_{hard_{i}}||_F\\
			&+  ||softplus( \max(\alpha_{i})-\min(\beta_{i}))||_{F})
		\end{aligned}
	\end{equation}
\end{small}
where  the subscript $F$ stands for frobenius norm, and $\gamma=0.2$ is  based on experiments on a hold-out validation set.

\subsubsection{Constructing  the  Text-vision Matrix (TVM)}
Based  on  the corrected \textbf{SSM} $\Theta$,  for the $k^{th}$ subregion, the \emph{dynamic representation} of words w.r.t. $h^{k}_{i-1}$ is	$q^{k}_{i-1}=\sum_{j=1}^{T}\theta_{j,k}w_j$. 
So, the  Text-Vision Matrix (TVM) for word  embeddings $W$  and image feature $H_{i-1}$  is denoted by  $Q_{i-1}=(q^{1}_{i-1}, q^{2}_{i-1}, ..., q^{N}_{i-1}) \in \mathbb{R}^{D \times N }$. The TVM  $Q_{i-1}$   and      image feature  $H_{i-1}$  are concatenated,  and   fed into   the   ResBlocks and   Upsampling modules to output the image feature  $H_i$.

\subsection{Layout Visual Refinement (LVR) loss}~\label{LVR_loss}

Based on the refined  layout structure,  we further enhance the visual representation within  the layout  area. 
To do this,  we  propose   Layout Visual Refinement (LVR) loss  to  enhance   the texture  perception and style information in the  layout.   
LVR loss  includes  Perception  Refinement (PR)  loss   and     Style  Refinement (SR) loss. 

\subsubsection{Perception  Refinement  loss}
To contruct   Perception  Refinement (PR) loss, we  first construct  the layout mask.    
In   \textbf{SSM}  $\Theta \in \mathbb{R}^{T \times N}$, each  column consists of   the semantics  similarity   weights between  one  subregion and  all  words.     
The maximum value  in  each  column     is  the  most related  word to the subregion.
Therefore, we use a maximization operation in each  column   of   $\Theta(^{*})$ to   get    $Mask_{\Theta(^*)}  \in  \mathbb{R}^{W_H \times H_H}$. 
Then we capture  the image feature within the layout: the layout  masks $Mask_{\Theta}$  and $Mask_{\Theta*}$  dot product   the  $H_i$ and  $H^{*}_i$,  respectively. 
Finally, the perception  refinement  loss  is  defined as 

\begin{equation}\label{PRL}
	L^{PR}_{i}=\frac{1}{N \cdot D} \|Mask_{\Theta_i} \odot H_i -Mask_{\Theta^{*}_i} \odot H^*_i \|_F
\end{equation} 
$L_{PR_i}$ can  drive  the  generator  to better  enhance   the texture  perception in the layout  area.

\subsubsection{ Style  Refinement loss}

To  further enhance  style-related information in the layout,  we align  the \textit{Gram  Matrix}   of  the   $Mask_{\Theta_i} \odot H_i$ with that of the   $Mask_{\Theta^{*}_i} \odot H^*_i$, and define the  Style  Refinement (SR) loss    as 

\begin{equation}~\label{SRL}
	L^{SR}_{i}= \frac{1}{N \cdot D} \| \mathcal{G}(Mask_{\Theta_i} \odot H_i) -\mathcal{G}(Mask_{\Theta^{*}_i} \odot H^*_i) \|_F
\end{equation}
where  $\mathcal{G}(\cdot)$ is  Gram Matrix calculation.

The LVR Loss   is  defined  as $L^{LVR}_{i}= \eta_1 \mathcal{L}^{PR}_{i}+   \eta_2 \mathcal{L}^{SR}_{i}$. The hyperparameters  $\eta_1=1.0$ and   $\eta_2=1.0$   are    based on experiments on a holdout validation set.

\subsection{Objective Functions  in ALR-GAN}~\label{Object_fucntions}
Combining the above modules, at the $i$-th stage of   ALR-GAN, the generative loss $\mathcal{L}_{G_i}$  is defined     as
\begin{footnotesize}
	\begin{equation}\label{eq10}
		L_{G_i} = \underbrace{-\frac{1}{2}\mathbb{E}_{\hat{I_i}\sim P_{G_i}}[logD_i(\hat{I_i})]}_{\text{unconditional loss}}-\underbrace{\frac{1}{2}\mathbb{E}_{\hat{I_i}\sim P_{G_i}}[logD_i(\hat{I_i},s)]}_{\text{conditional loss}},
	\end{equation}
\end{footnotesize}
where the unconditional loss   pushes the   synthesized  image to  be more realistic,   to fool  the discriminator, and  the conditional loss  drives   the  synthesized  image   to  better match  the   corresponding text description.  The  discriminative  loss   is defined as 

\begin{scriptsize}
	\begin{equation}\label{eq11}
		\begin{aligned}
			L_{D_i} = &\underbrace{-\frac{1}{2}\mathbb{E}_{I^*_i\sim P_{data_i}}[logD_i(I^*_i)] - \frac{1}{2}\mathbb{E}_{\hat{I_i}\sim P_{G_i}}[log(1-D_i(\hat{I_i})]}_{\text{unconditional loss}}+ \\
			&\underbrace{-\frac{1}{2}\mathbb{E}_{I^*_i\sim P_{data_i}}[logD_i(I^*_i,s)] - \frac{1}{2}\mathbb{E}_{\hat{I_i}\sim P_{G_i}}[log(1-D_i(\hat{I_i},s)]}_{\text{conditional loss}},
		\end{aligned}
	\end{equation}
\end{scriptsize}
where $I^*_{i}$  and  $\hat{I_i}$  are the   $i$-th scale  image,  the  discriminative  loss $\mathcal{L}_{D_i}$   classifys the  input image sampling  from  the real  image  distribution or  synthesized  image  distribution.

To generate realistic images, the final objective functions of the  generative   and discriminative networks are defined  respectively as

\begin{small}
	\begin{equation}\label{eq121}
		L_{G}=\sum_{i=0}^{m-1}L_{G_i} +  \sum_{i=1}^{m-1}[L^{ALR}_{i}+\lambda_1 L^{REC}_{i}+  L^{LVR}_{i}]+\lambda_2 L_{DAMSM},
	\end{equation}
\end{small}
\begin{small}
	\begin{equation}\label{eq12}
		L_{D}= \sum_{i=0}^{m-1}L_{D_i}.		
	\end{equation}
\end{small}
Here,  ALR-GAN has three stage generators ($m=3$), and $\lambda_1=10^{-1}$. 
The parameter  $\lambda_2=50$ in the MS-COCO dataset,  and the parameter  $\lambda_2=5$ in the CUB-Bird dataset. The   values   of  $\lambda_2$   in ALR-GAN  are the same as in AttnGAN~\cite{Xu2017AttnGAN}  and  DM-GAN \cite{Minfeng2019}.  The parameter  $\lambda_2$  is  used  to  balance  the   loss  $\mathcal{L}_{DAMSM}$ in the generative loss $L_G$, and is  introduced  to  improve  the semantics-matching  between    image subregions and words. 

\begin{table*}[]
	\centering
	\caption{ \textbf{IS} $\uparrow$,  \textbf{FID} $\downarrow$, \textbf{SOA-C/SOA-I} $\uparrow$,  and \textbf{R-Precision} $\uparrow$ by some SOTA GAN-based T2I models and   ALR-GAN on the CUB-Bird and MS-COCO  test sets.  \textbf{W A.I.: With Auxiliary Information. W/O A.I.: Without Auxiliary Information.} Scores for models marked with * were calculated with  pretrained models provided by the respective authors.  First, second, and third  scores  are shown  in  \textcolor{red}{red}, \textcolor{ForestGreen}{green}, and \textcolor{blue}{blue}, respectively.}
	\setlength{\tabcolsep}{1.5mm}{
		\begin{tabular}{c c|c c c | c  c  c  c   c }
			\toprule[1.15pt]
			\multirow{2}{*}{Pattern}   & \multirow{2}{*}{Method}  	&		
			\multicolumn{3}{|c|}{CUB-Bird} &\multicolumn{5}{|c}{MS-COCO}  \\
			\cline{3-10} 	
			& 	& IS      &  FID  	& R-Presion  (\%)      &  IS  	& FID      &  R-Presion (\%)          & SOA-C  & SOA-I   \\	
			\hline
			&	Obj-GAN \cite{Wenb19cvpr}     & -  & -  & -  &  $30.29 \pm 0.33$  & $25.64$ & $91.05$  &  $27.14^*$   &  $41.24^*$  \\ 
			W A.I. &	OP+AttnGAN~\cite{Semantic_c_2020}  & - & - & - &  $27.88 \pm 0.12$ & $24.70$   &  $89.01$ &   $35.85$ & $50.47$ \\ 
			&	CP-GAN~\cite{Liang2020CPGANCG}    & - & - & - &  $52.73 \pm 0.61$ & -   &  $93.59$ &   $77.02$ & $84.55$ \\ 
			&	RiFe-GAN~\cite{Juncvpr2020}    & $5.23 \pm 0.09$  & - & $\sim22.5$(Average)  & $31.70$ &  - &   - & - \\   \hline     \hline 			                      
			&	AttnGAN \cite{Xu2017AttnGAN}     &  $4.36 \pm  0.02$    & $23.98$     &   $52.62^*$   &  $25.89 \pm 0.19$ & $35.49$ & $61.34^*$   &  $25.88^*$ &  $39.01^*$ 	\\ 
			&	CSM-GAN~\cite{9359527}	 &  $4.62 \pm 0.08$   & $20.18$       & $54.92^*$    &  $26.77 \pm 0.24$ & $33.48$ &  $63.84^*$    &  $27.24^*$ &  $42.10^*$  \\
			&	SE-GAN \cite{thc2019}     & $4.67 \pm 0.04$ & $18.17$ &  -     & $27.86  \pm  0.31$  &  $32.28$  &  - & - & -   \\	   
			
			&	DM-GAN	\cite{Minfeng2019}  &  $4.75 \pm 0.07$   &  $16.09$  &  $59.21^*$     &  $30.49 \pm 0.57$ & $32.64$   &  $70.63^*$  & $33.44^* $  &  $48.03^*$   \\ 	    
			W/O A.I.   &	KT-GAN~\cite{9210842}   &  $4.85 \pm 0.04$ & $17.32$ &   \textcolor{ForestGreen}{$70.76^*$}     &  $31.67 \pm 0.36$  & $ 30.73$  & $65.20^*$  & $35.02^*$  & $42.86^*$  \\                      
			& DF-GAN~\cite{DBLP-DF-GAN} &  $4.86 \pm 0.04$   &  $19.24$   & -   & -  & -  & -   & -  & -  \\ 
			& DAE-GAN~\cite{Ruan2021DAEGANDA}   &  $4.42  \pm  0.04$  & $15.19$  &  $59.44^*$   &  $35.08 \pm 1.16$  &  $28.12$ & $73.81^*$ &  $38.41^*$ & $57.30^*$\\
			& DR-GAN~\cite{9760725}   &  \textcolor{blue}{$4.90 \pm 0.05$}    & \textcolor{blue}{$14.96$}  &  $61.17^*$   &  $34.59 \pm 0.51$  &  $27.80$ & \textcolor{blue}{$77.27^*$} &  $41.62^*$ & $60.73^*$  \\  
			&	XMC-GAN \cite{Zhang2021CrossModalCL}     & -  & -  & -     & $30.45$  &  \textcolor{red}{$9.33$}  &  $71.00$  &  \textcolor{ForestGreen}{$50.94$}  &  \textcolor{blue}{$71.33$} \\\hline  \hline 
			
			& Baseline   &  $4.51 \pm 0.04$   &  $23.32$    & $71.63$  &   $29.17  \pm 0.22$    &  $33.67$    &  $61.29$   &   $31.74$    &  $47.12$ \\
			& ALR-GAN  &   \textcolor{red}{4.96 $\pm$ 0.04}   & $15.14$   & \textcolor{red}{$77.54$}    &  $34.70  \pm  0.66$    & $29.04$    & $69.20$    &  $42.47$  &  $62.20$    \\	
			W/O A.I.   & DMGAN~\cite{Minfeng2019}+ALR+LVR   &  $4.81 \pm 0.06$ &  $15.37$ &  $63.90$   &  \textcolor{ForestGreen}{$35.27  \pm  0.46$} &  $29.42$    &  $74.28$  &  $43.13 $  &  $58.90$   \\			
			& DAE-GAN~\cite{Ruan2021DAEGANDA}+ALR+LVR  &  $4.53  \pm  0.09$  &  \textcolor{red}{$14.26$}   &  $62.03$   &  \textcolor{red}{$37.02 \pm 0.76$}  &  \textcolor{ForestGreen}{$23.79$}   &  \textcolor{red}{$79.40$}    &  \textcolor{red}{$55.64$}  & \textcolor{ForestGreen}{$72.56$} \\
			& DR-GAN~\cite{9760725}+ALR+LVR  &  \textcolor{ForestGreen}{$4.94\pm  0.03$}  & \textcolor{ForestGreen}{$14.70$}  &  \textcolor{blue}{$64.40$}   &  \textcolor{blue}{$35.12 \pm 0.57$} & \textcolor{blue}{$26.03$} &  \textcolor{ForestGreen}{$79.04$} &  \textcolor{blue}{$49.73$}  & \textcolor{red}{$73.10$}\\

			\bottomrule[1.15pt]	
			
	\end{tabular} }
	\label{SOTA-Results} 
\end{table*}

\section{Experimental  Results}~\label{subsection4}

We discuss  the experiment settings,   compare   ALR-GAN  with many GAN-based T2I methods, and    evaluate the  effectiveness of each component. In the training stage: (i) in the CUB-Bird dataset, the  training   batch size  is $16$, and the  training epoch is  $800$; (2) in the MS-COCO dataset, the  training   batch size  is $14$, and  the  training epoch is  $150$.  All experiments about ALR-GAN are trained and tested on one GTX 3090 GPU respectively.

\subsection{Experiment Settings}

\subsubsection{Datasets}
ALR-GAN was evaluated on  two   public  datasets,  CUB-Bird~\cite{WahCUB_200_2011} and  MS-COCO~\cite{Lin2014Microsoft}.  
CUB-Bird   contains $11,788$ bird images, and $10$ sentences for each image. 
MS-COCO   contains $80$K training images and $40$K testing  images,   each    with  five  sentences. 
The testing   and training sets  were preprocessed using the same pattern  as  in~\cite{Reed2016Generative, Han2017StackGAN}. 
We  adopt  the image generated by the first sentence in the testing  dataset as our  ``Testing  Set''.  
The data  setting is the same as these  SOTA T2I methods~\cite{Minfeng2019, Ruan2021DAEGANDA, 9760725}.
Besides, we adopt the image generated by the second sentence in the testing  dataset as  our  ``Validation Set''.
So,  the number of the ``Testing set'' and ``Validation set'' in CUB-Bird is $2933$;  the number of the  ``Testing set'' and ``Validation set''  in MS-COCO is $\sim$ $40K$.

\subsubsection{Baseline}
Our  baseline model is   AttnGAN~\cite{Xu2017AttnGAN}   with   Spectral Normalization~\cite{Normalization2018}, which limits   drastic gradient changes and improves   training efficiency, and hence model performance.   
Due to GPU memory limitations,  the Baseline model  includes three generators,   each with a corresponding discriminator.  
The first through third generators generate $64 \times 64$, $128 \times 128$, and $256 \times 256$ images, respectively.

\subsubsection{Evaluation} 
We use  four measures  to  evaluate  the  performance  of    ALR-GAN, where $\uparrow$ means that the higher the value the better the performance, and vice versa.  
\textbf{Inception Score (IS$\uparrow$): }~\cite{Salimans2016ImprovedTF} This is obtained by   an Inception-V3 model   fine-tuned by~\cite{Han2017StackGAN}  to calculate  the KL-divergence between the marginal and conditional class distributions. 
A large IS indicates  that synthesized  images have  high diversity for all classes, and each image  can  be recognized as a specific class. 
\textbf{Fr\'{e}chet Inception Distance (FID$\downarrow$):}~\cite{Heusel2017GANsTB}  A lower FID score  between    synthesized    and   real images  means that the synthesized  image distribution is closer to the real image distribution,  and    that the generator  can  synthesize photo-realistic images.
\textbf{Semantic Object Accuracy (SOA$\uparrow$  ):} This includes  	SOA-C     and  SOA-I scores.  	SOA-C      is  the percentage of synthesized images per class in which the	desired object is detected, and SOA-I   is  the percentage	of synthesized  images in which the desired object is detected. 
\textbf{R-precision$\uparrow$:}  This is used  to evaluate the semantic consistency   between a   synthesized  image and the corresponding  text description.

\subsection{Comparison with state-of-the-art GAN models}

\textbf{Image Diversity and Objective Evaluation.}
We use   IS   scores, as shown  in Table~\ref{SOTA-Results}, to evaluate  the  objectives and diversity  of  synthesized images  on the CUB-Bird and   MS-COCO test sets.  
Compared with  T2I methods Without  Auxiliary Information  (W/O A.I. in Table~\ref{SOTA-Results}),     ALR-GAN  performs     competitively. 
The IS score  of  DAE-GAN~\cite{Ruan2021DAEGANDA}   is higher than  that  of   ALR-GAN on the MS-COCO  dataset because DAE-GAN uses extra NLTK POS tagging and manually designs rules for different datasets. 
Compared with  T2I methods With  Auxiliary Information  (W A.I. in Table~\ref{SOTA-Results}),  the IS score  of  RiFe-GAN~\cite{Juncvpr2020}  is higher than  that  of   ALR-GAN on   CUB-Bird, and  the IS score  of  CP-GAN~\cite{Liang2020CPGANCG}  is higher than  that  of   ALR-GAN on MS-COCO, because RiFe-GAN~\cite{Juncvpr2020} requires additional  text sentences to train the model. 
CP-GAN~\cite{Liang2020CPGANCG}  requires  additional auxiliary information, including the  object’s bounding box and shape, to train the T2I model.  
In contrast,  we only explore  the layout  information from  the   image  and  the  text description, without      additional auxiliary information,   to guide  the generator to correct  the  layout, to synthesize a high-quality  image.  Here,  the DAE-GAN+ALR+LVR and DR-GAN+ALR+LVR are   trained on one  Tesla V100 GPU from scratch.  We adopt the DAE-GAN and DR-GAN pre-trained model released by  authors to synthesize images on one GTX 3090 GPU respectively.  All experiments about DM-GAN+ALR+LVR  are   trained and tested on one  GTX 3090 GPU. We also train the   DM-GAN+ALR+LVR   from scratch.

\begin{figure*}[h!tb]
	\centering
	\includegraphics[scale=0.235]{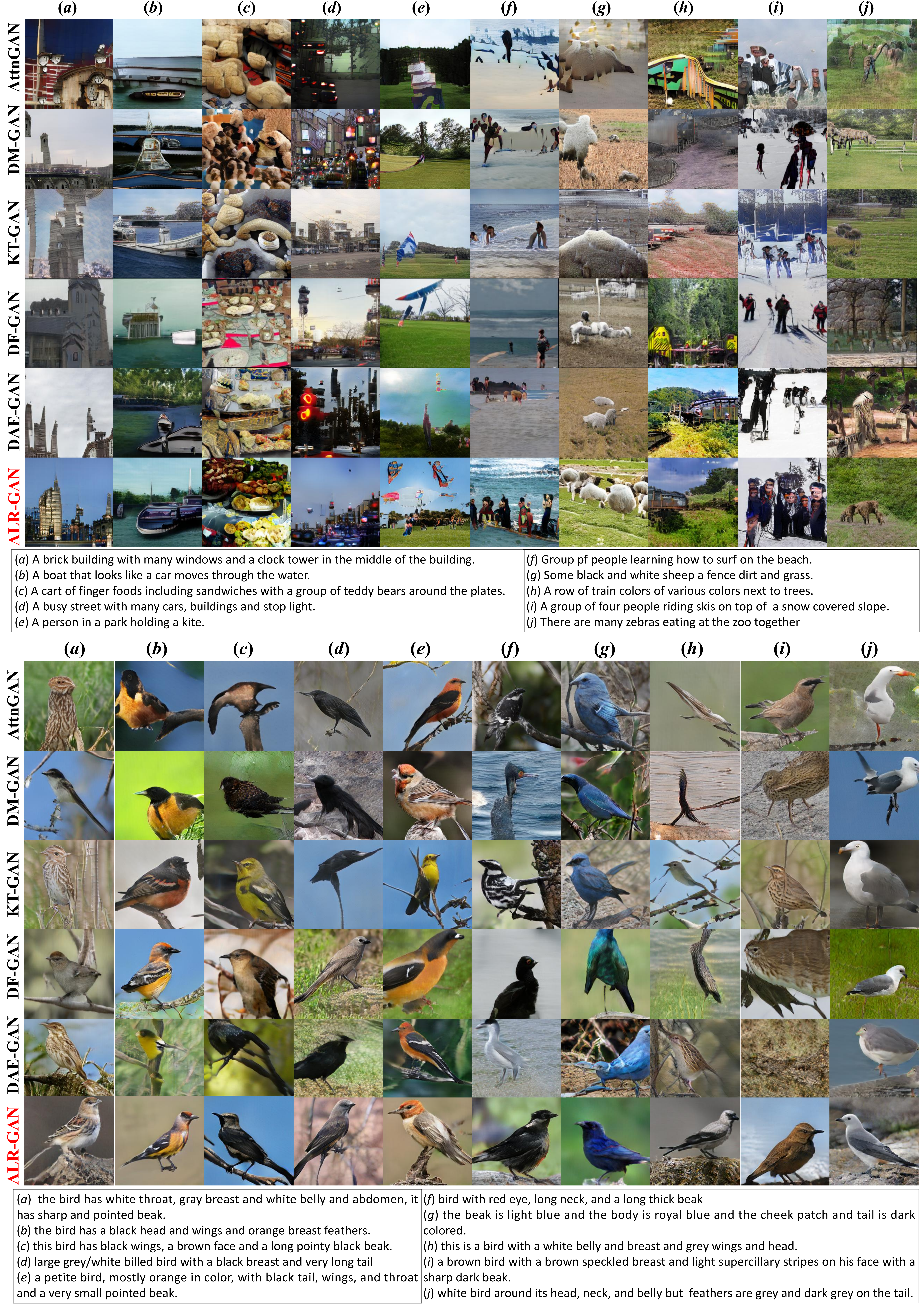}
	\caption{Images of $256\times256$ resolution are synthesized by AttnGAN, DM-GAN, KT-GAN, DF-GAN, DAE-GAN,  and   ALR-GAN  conditioned on text descriptions from   MS-COCO  dataset (Upper Part) and CUB-Bird dataset (Lower Part).} 
	\label{SOTA_COCO_Image1} 
\end{figure*}

\textbf{Semantic Object Accuracy Evaluation.}   
We  adopt the  SOA score (which includes the SOA-C     and SOA-I scores)  \cite{Semantic_c_2020}  to  evaluate the quality of individual  objects and  regions  within an image on  the MS-COCO  dataset. 
As  shown in Table~\ref{SOTA-Results},  the SOA-C    and SOA-I scores of     ALR-GAN  are  much  better  than those of most   T2I methods (including DAE-GAN~\cite{Ruan2021DAEGANDA})  and Baselines  without  auxiliary information.
ALR-GAN is better than    methods with auxiliary information, such as Obj-GAN \cite{Wenb19cvpr} and OP-GAN~\cite{Semantic_c_2020}.  
The  SOA-C    and SOA-I scores of XMC-GAN \cite{Zhang2021CrossModalCL} and CP-GAN~\cite{Liang2020CPGANCG}   are higher than those of    ALR-GAN. 
 CP-GAN~\cite{Liang2020CPGANCG} requires additional auxiliary information,  including the object’s bounding box and shape.   
XMC-GAN  adopted contrastive learning  to   capture inter-modality and  intra-modality correspondences, which enriched  the global semantics and region semantics  of the synthesized image.  The  intra-modality correspondence can  push the synthesized  image into the real image.   The synthesized  image distribution is closer to the real image distribution. 
The SOA-C  is  the percentage of synthesized images per class in which the	desired object is detected, and SOA-I   is  the percentage	of synthesized  images in which the desired object is detected.   Sufficient global and regional semantics of synthesized images are conducive to object  recognition, so XMC-GAN gets  high  SOA-I/SOA-C  scores.

\textbf{Distribution and  Semantics  Consistency Evaluation.} 
We  use  the FID  to evaluate  distribution  consistency  between    real    and    synthesized  images.
We  use  the R-precision proposed  by  AttnGAN~\cite{Xu2017AttnGAN}  to evaluate  the  semantic  consistency  between   the  text  description  and     synthesized  image.
As shown  in  Table~\ref{SOTA-Results},      ALR-GAN  achieves    competitive  performance on   two measures, and  is much  better  than  our Baseline.  
The FID score  of DAE-GAN~\cite{Ruan2021DAEGANDA} is higher than  that  of    ALR-GAN  because DAE-GAN uses extra NLTK POS tagging and manually-designed rules for different datasets, while we only explore  the layout  information from  the   image  and  the  text description, without      additional auxiliary information.

\begin{figure}[h!tb]
	\centering
	\includegraphics[scale=0.115]{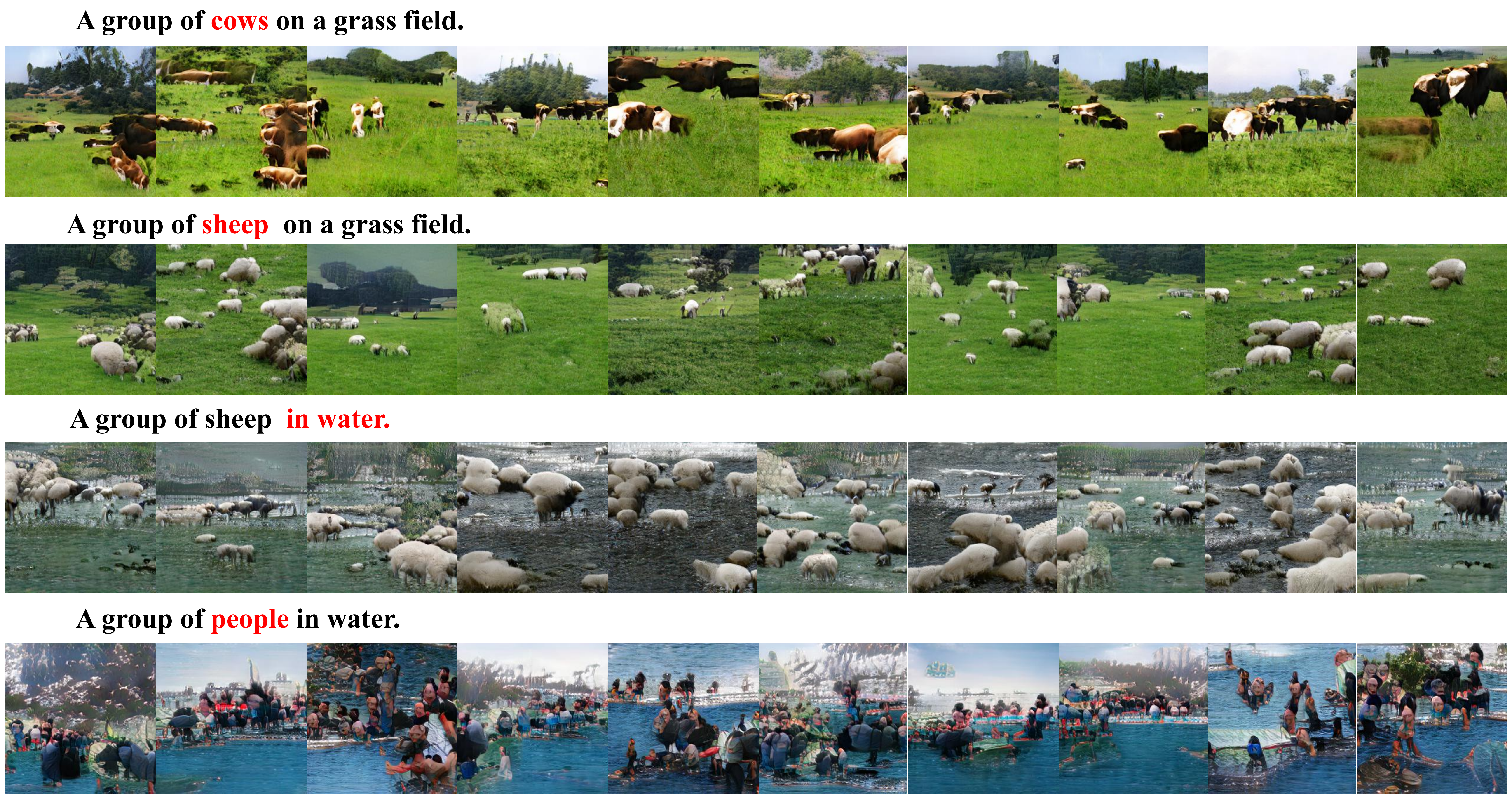}
	\caption{Examples of ALR-GAN on   ability to catch subtle
		changes (phrase or words in red) of   text descriptions
		on  MS-COCO  test set, and  synthesizing  diverse images with    reasonable layout.} 
	\label{Diversity_Figure} 
\end{figure}

\textbf{Generalization.}
To observe  the  generalization  of  the  ALR  module and    LVR  loss,  we     introduce   them  into  DMGAN~\cite{Minfeng2019}, DR-GAN~\cite{9760725}, and  DAE-GAN~\cite{Ruan2021DAEGANDA}.  
As shown in Table~\ref{SOTA-Results},  these  two    designs  can help  these    methods   perform better  on  different  measures  because they  are  designed to  refine  the layout structure and   enhance the visual representation of the object or background within the layout.
However,  these   components  intuitively constrain the diversity of  layouts. 
In fact,  the layouts  of  objects  in training samples are diverse. 
Thus, they  guide the generator  to better learn the distribution of the object's layout. 
Therefore, for the same text description, the generator  still  can synthesize  diverse  layouts, as shown in Fig.~\ref{Diversity_Figure}.
Hence our proposed ALR module and LVR loss are plugin designs that can be applied to other T2I models.

\begin{figure*}[h!tb]
	\centering
	\includegraphics[scale=0.29]{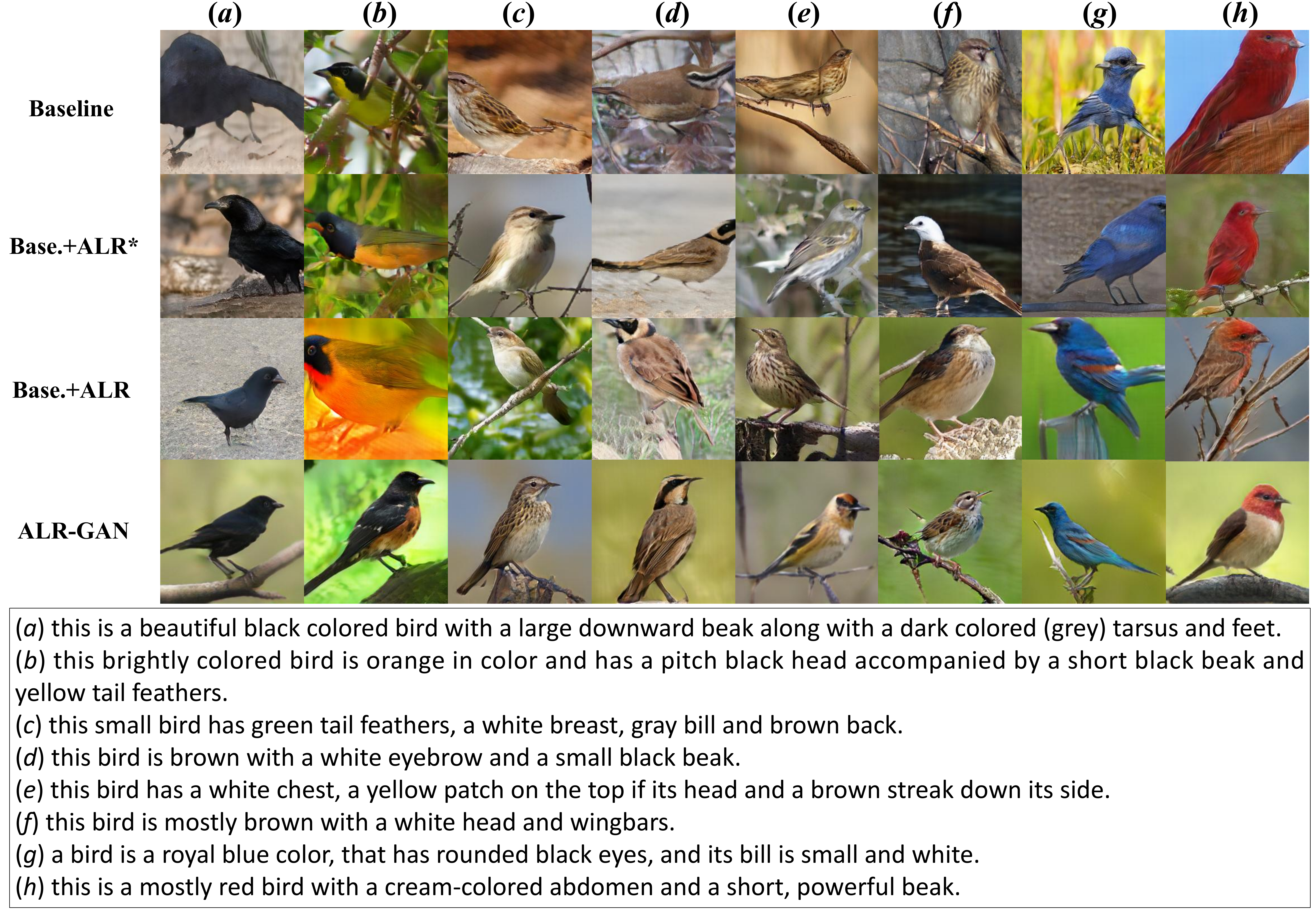}
	\caption{Images ($256\times256$)    synthesized by   Baseline (Base.), Base.+ALR,  and ALR-GAN   conditioned on text descriptions from   CUB-Bird  test set. } 
	\label{Ablation_Image} 
\end{figure*}

\begin{figure*}[h!tb]
	\centering
	\includegraphics[scale=0.62]{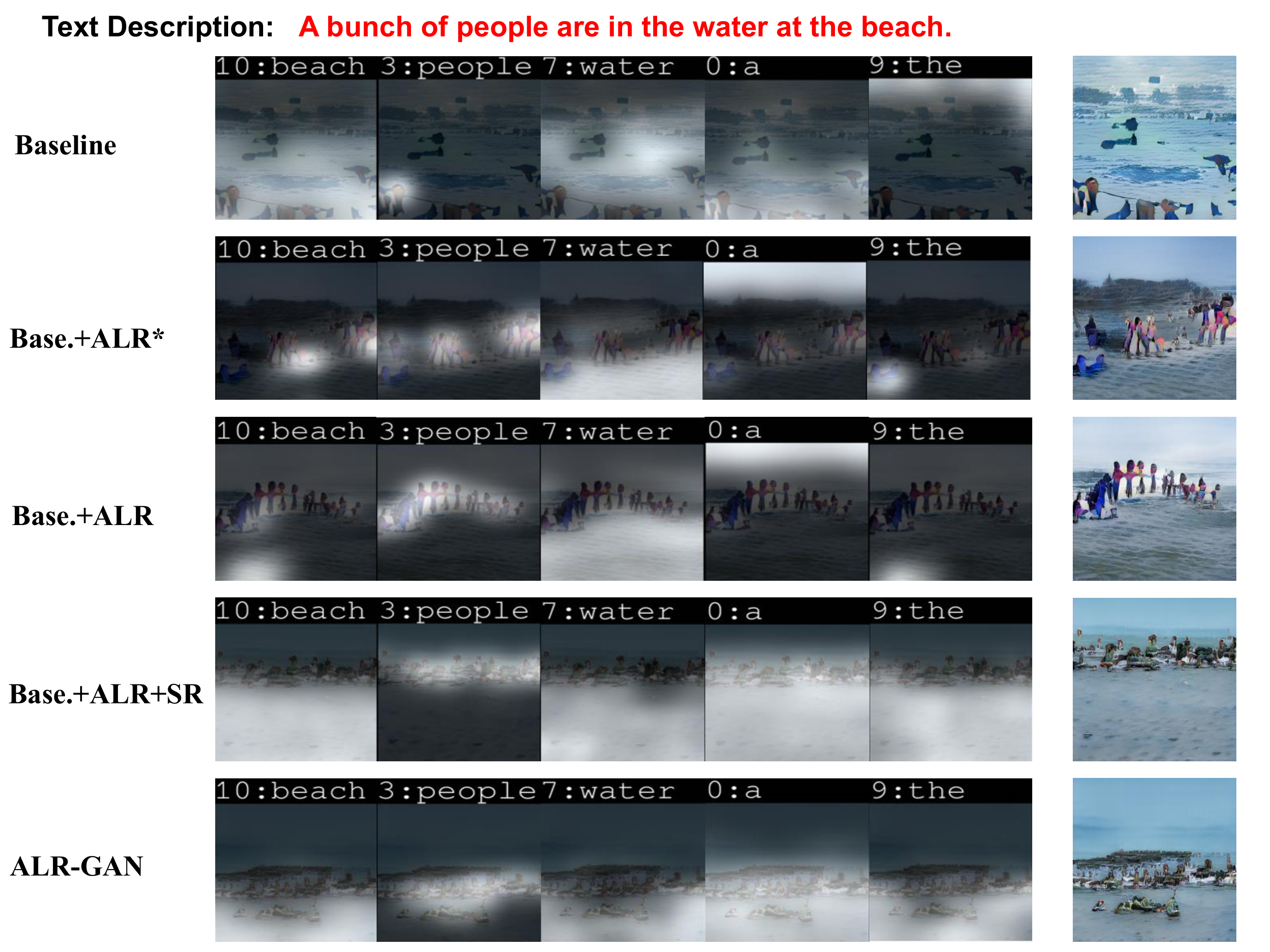}
	\caption{Semantics-matched  regions  between   words and   image from Baseline (Base.), Base.+ALR$^*$, Base.+ALR,  Base.+ALR+PR,   and   ALR-GAN. Highlighted region is   semantics-matched  region  between   word and    whole image.} 
	\label{Layout_Ab_COCO} 
\end{figure*}

\textbf{Visualization Evaluation.} 
We qualitatively  evaluate    ALR-GAN and some  outstanding  GAN-based  T2I  methods  by   image  visualization on the  CUB-Bird and  MS-COCO  test sets. 
On CUB-Bird, the image is dominated by a single bird.  
Compared with  other methods,     ALR-GAN can improve   the layout and detail semantics of various parts of the bird's body in  the Lower Part  of Fig.~\ref{SOTA_COCO_Image1}. 
On  MS-COCO, images are usually scene-type images with a variety of objects or backgrounds.  
Compared with   CUB-Bird, there is little specific appearance information   of a single object   in MS-COCO. 
Therefore, the visual quality of the synthesized  objects in  this dataset is not  as good as   in    CUB-Bird.  
We can observe from  the Upper Part  of Fig.~\ref{SOTA_COCO_Image1} that the layout  quality   synthesized by ALR-GAN is more reasonable than by other T2I	 methods. Although DAE-GAN has a better IS score on COCO  than ALR-GAN, the latter focuses on layout generation. On the COCO dataset,  the layout synthesized by ALR-GAN  is better than by DAE-GAN.

We further evaluate the sensitivity of the proposed ALR-GAN by changing just one word or phrase in the input sentence.  
As shown in Fig.~\ref{Diversity_Figure}, the synthesized  images are modified according to the words or phrase changes of the input text  description, e.g.,  image scene and  objects (``in  water,'' ``on a  grass field,''  ``people,'' ``cows,'' ``sheep''). 
We  can see  that   ALR-GAN also can  catch subtle changes of the text  description  and  retain   semantic diversity from text  and reasonable  layouts.    

\textbf{Model Cost.}
Table~\ref{model_size}  compares
ALR-GAN  with    SOTA T2I  methods  under four model cost  measures:  Training Time, Training Epoch,  Model  Size,  and  Testing Time.  Using the MS-COCO dataset as an example, compared with these methods, the model cost of   ALR-GAN is appropriate.

\begin{table}[h]
	\centering
	\caption{   Training Time, Training Epochs,  Model  Size, and Testing Time    of   ALR-GAN and  SOTA T2I methods   on    MS-COCO dataset. }	
	\setlength{\tabcolsep}{0.1mm}{
		\begin{tabular}{c|c |c | c | c }
			\toprule[1.15pt]
			$\textbf{Method}$     & Training Time  &    Training Epoch  & Model  Size  & Testing Time   \\   \hline
			AttnGAN~\cite{Xu2017AttnGAN}    & $\sim 8 Days$  &  $120$ &  $\sim 55.5M$ &  $\sim 1200s$  \\  
			DM-GAN\cite{Minfeng2019}     & $\sim 14 Days$  &  $200$ &  $\sim 89.7M$ &  $\sim 1800s$  \\  
			DAE-GAN\cite{Ruan2021DAEGANDA}     & $\sim 11 Days$  &  $200$ &  $\sim 91M$ &  $\sim 1900s$  \\  
			DF-GAN\cite{DBLP-DF-GAN}     & -  &  $300$ &  $\sim 189M$ &  -  \\ 	
			DR-GAN\cite{9760725}     & $\sim 10 Days$ & $150$ & $\sim 73.2M$ & $\sim 1400s$  \\ 						
			ALR-GAN(Ours)  &  $\sim 10 Days$ & $150$ & $\sim 76M$ & $\sim 1400s$   \\  
			\bottomrule[1.15pt]		
	\end{tabular} }
	\label{model_size}  	
\end{table}

\subsection{Ablation Study}

We evaluate the effectiveness of  the  ALR  module   and    LVR loss.
The  IS score, FID score,   SOA-C score,   and SOA-I score are shown in Table~\ref{ablation_table}.  The IS     and FID   scores  produced by combining different components on  the CUB-Bird dataset; the  SOA-C score and SOA-I  score  produced by combining different components on  the MS-COCO dataset.

We gradually add these components to the Baseline (Base.), and introduce  the  ALR module, i.e., Base.+ALR.  
As shown in Table~\ref{ablation_table},  compared  with the  Baseline,  on the CUB-Bird dataset, Base.+ALR  improves  the IS score  from  $4.51$  to $4.79$, and  the FID score  of Base.+ALR  drops from $23.32$ to $17.33$; on the MS-COCO Dataset, Base.+ALR  improves  the SOA-C score  from  $31.74$  to $38.12$, and improves  the SOA-I score  from  $47.12$  to $55.81$.   
We   introduce  the Perception  Refinement  (PR) loss   to  Base.+ALR,  i.e.,  Base.+ALR+PR. 
As shown in Table~\ref{ablation_table},  compared  with    Base.+ALR,  on the CUB-Bird dataset,  Base.+ALR+PR  improves  the IS score  from  $4.79$  to $4.84$, and the FID score  of Base.+ALR+PR  drops from $17.33$ to $16.08$;  on the MS-COCO Dataset, Base.+ALR+PR   improves  the SOA-C score  from  to $38.12$   to  $40.28$, and improves  the SOA-I score  from   $55.81$  to $57.29$.   
We  introduce  the LVR loss  to   Base.+ALR, i.e.,    ALR-GAN. 
Compared with    Base.+ALR+PR,   ALR-GAN     improves  the  performance  over  four  measures. Note that the LVR loss   contains  the Perception  Refinement  (PR)     and     Style  Refinement (SR) loss.

\begin{table}[]
	\centering
	\caption{  IS $\uparrow$  and FID $\downarrow$    produced by combining different components on    CUB-Bird dataset.     SOA-C  $\uparrow$  and SOA-I $\uparrow$  produced by combining different components on    MS-COCO dataset. }
	\setlength{\tabcolsep}{2mm}{
		\begin{tabular}{c|c c | c c}
			\toprule[1.15pt]
			\multirow{2}{*}{Method}   &
			
			\multicolumn{2}{|c|}{CUB-Bird} &\multicolumn{2}{|c}{MS-COCO}  \\
			\cline{2-5} 	
			& IS      &  FID         & SOA-C & SOA-I   \\	
			\hline
			Baseline (Base.) & 4.51 $\pm$ 0.04  & $23.32$    & $31.74$  &  $47.12$   \\   \hline  \hline 
			Base.+ALR  & 4.79$\pm$ 0.03 & $17.33$  & $38.12$  &  $55.81$      \\  
			Base.+ALR+PR  & 4.84 $\pm$ 0.05  & $16.08$   & $40.28$  &  $57.29$     \\
			ALR-GAN  & 4.96 $\pm$ 0.04  &     $15.14$   & $42.47$  &  $62.20$    \\  \hline \hline 
			Base.+ALR$^*$   &  4.58$\pm$ 0.03  & $21.03$   & $35.85$  &  $50.11$    \\ 
			Base.+ALR+PR$^*$ &  4.81$\pm$ 0.02  & $16.92$   & $39.65$  &  $56.31$    \\ 
			Base.+ALR+PR+SR$^*$   &  4.86$\pm$ 0.07  & $16.13$  & $41.63$  &  $59.44$      \\ 
			\bottomrule[1.15pt]	
			
	\end{tabular} }
	\label{ablation_table} 
\end{table}

We  discuss  the effectiveness of  the   adaptive weight adjustment mechanism   in $L^{ALR}_i$ (Eq.~\ref{AACL_Eq}). 
We   substitute   $\| \Theta -\Theta^{*} \|_F$  for $L^{ALR}_i$ (Eq.~\ref{AACL_Eq}) in the ALR module, i.e., Base.+ALR$^\dagger$,  in  Table~\ref{ablation_table}. 
Base.+ALR$^*$ performs worse  than   Base.+ALR. As  shown in Fig.~\ref{Ablation_Image} and Fig.~\ref{Layout_Ab_COCO},  the  layout  structure  of the synthesized image  is  worse than  that  of Base.+ALR.  
This indicates that   the  adaptive weight adjustment mechanism can effectively improve the matching efficiency between $\Theta$ and  $\Theta^{*}$, to further improve image  quality.

We   discuss  the importance  of     Perception  Refinement  (PR) loss  $L^{PR}_i$ (Eq.~\ref{PRL})   in LVR loss $L^{LVR}_i$.  
We    substitute   $\frac{1}{N \cdot D} \| H-H^{*} \|_F$  for  $L^{PR}_i$ (Eq.~\ref{PRL}), i.e.,  Base.+ALR+PR$^*$,  in  Table~\ref{ablation_table}. 
Compared  with  Base.+ALR+PR,  the performance  of    Base.+ALR+PR$^*$ is    worse. 
We  further discuss  the importance  of    Style  Refinement (SR) loss ($L^{SR}_i$, Eq.~\ref{SRL})  in LVR loss $L^{LVR}_i$.   As shown in Fig.~\ref{Layout_Ab_COCO},  from Base.+ALR   to Base.+ALR+SR,  the style    and   layout information  of  the synthesized  images is significantly improved.  We   further   substitute   $\frac{1}{N \cdot D} \| \mathcal{G}(H) -\mathcal{G}(H^*) \|_F$  for  $L^{SR}_i$ (Eq.~\ref{SRL}),  i.e.,  Base.+ALR+PR+SR$^*$, in  Table~\ref{ablation_table}.   
Compared  with  ALR-GAN,   the performance  of    Base.+ALR+PR+SR$^*$ is  also worse. 
Through the experimental results, we can simply analyze the reasons.
This  is  because the semantics of the text description   only covers part of the semantics of the image. 
Over-constraint on the layout and details that are not included in the text description can significantly increase the training burden of the model. 
For the  synthesized image, this additional uncontrollable visual information generation will interfere with image quality evaluation.

Finally, we     qualitatively evaluate the  effectiveness  of each  component by image visualization (Fig.~\ref{Ablation_Image} and Fig.~\ref{Layout_Ab_COCO}).  
In  Fig.~\ref{Ablation_Image},  we can see that  these proposed  strategies   can  effectively  improve  the details, structure, and layout  of birds. 
And, with the addition of losses, the structure and layout  of  the  birds    gradually improves.
In Fig.~\ref{Layout_Ab_COCO}, we can also see that with the introduction of modules or loss functions, the attention area   and  layout  structure of  the scene, the style information are gradually improved.
Overall, these  ablation  studies indicate  that  the ALR  module  and    LVR loss     can effectively  improve  the details, structure, and layout  of the  synthesized  images.

\subsection{Parametric Sensitivity Analysis}
We analyze  the   sensitivity of   \textbf{$\gamma$, $\eta_1$, $\eta_2$, $m$, and $\lambda_1$}.   

\begin{table}[]
	\centering
	\caption{IS $\uparrow$  and FID $\downarrow$    of  different  values  of      $\gamma$  on CUB-Bird dataset.   SOA-C  $\uparrow$  and SOA-I $\uparrow$  of  different  values  of    $\gamma$   on    MS-COCO dataset. }
	\setlength{\tabcolsep}{3mm}{
		\begin{tabular}{c|c c | c c}
			\toprule[1.15pt]
			\multirow{2}{*}{Method}   &
			
			\multicolumn{2}{|c|}{CUB-Bird} &\multicolumn{2}{|c}{MS-COCO}  \\
			\cline{2-5} 	
			& IS      &  FID         & SOA-C & SOA-I   \\	
			\hline
			
			$\gamma=0, 1.0$  & $\sim 4.6$ &   $\sim 20.1$  & $\sim 36.4$  &  $\sim 52.2$      \\ 
			
			$\gamma=0.1$  & 4.89$\pm$ 0.04 &    $16.11$  & $41.87$  &  $58.61$      \\

			$\gamma=0.2$ & 4.96 $\pm$ 0.04  &     $15.14$   & $42.47$  &  $62.20$     \\	
			$\gamma=0.3$  & 4.94 $\pm$ 0.07 &    $15.59$  &  $42.65$   & $60.22$    \\ 
			
			$\gamma=0.5$  & 4.90 $\pm$ 0.04 &    $15.64$  &  $42.66$   & $59.96$    \\ 	
			
			$\gamma=0.8$  & 4.83 $\pm$ 0.03 &    $16.22$  &  $39.45$   & $54.73$    \\
			
			\bottomrule[1.15pt]	
			
	\end{tabular} }
	\label{ablation_parameter} 
\end{table}

\begin{table}[]
	\centering
	\caption{IS $\uparrow$  and FID $\downarrow$  under  different numbers of generators  on  CUB-Bird dataset.   SOA-C  $\uparrow$  and SOA-I $\uparrow$ are under  different numbers of generators  on    MS-COCO dataset.   $m=0, 1, 2$:  number of generators is $1$, $2$, and $3$, respectively. }
	\setlength{\tabcolsep}{3mm}{
		\begin{tabular}{c|c c | c c}
			\toprule[1.15pt]
			\multirow{2}{*}{Method}   &
			
			\multicolumn{2}{|c|}{CUB-Bird} &\multicolumn{2}{|c}{MS-COCO}  \\
			\cline{2-5} 	
			& IS      &  FID         & SOA-C & SOA-I   \\	
			\hline
			
			$m=0$  &  2.91 $\pm$ 0.10    &   $48.22$   & $5.27$  &  $15.03$    \\ 			
			$m=1$  &  4.05 $\pm$ 0.08 &     $32.17$   & $22.53$  &  $36.74$      \\ 			
			$m=2$ & 4.96 $\pm$ 0.04  &     $15.14$   & $42.47$  &  $62.20$     \\				
			\bottomrule[1.15pt]	
			
	\end{tabular} }
	\label{ablation_parameter1} 
\end{table}

\begin{table}[]
	\centering
	\caption{IS $\uparrow$  and FID $\downarrow$    of  different values of        $\eta_1$  on   CUB-Bird dataset.    SOA-C  $\uparrow$  and SOA-I $\uparrow$  of  different values of       $\eta_1$   on    MS-COCO dataset. }
	\setlength{\tabcolsep}{3mm}{
		\begin{tabular}{c|c c | c c}
			\toprule[1.15pt]
			\multirow{2}{*}{Method}   &
			
			\multicolumn{2}{|c|}{CUB-Bird} &\multicolumn{2}{|c}{MS-COCO}  \\
			\cline{2-5} 	
			& IS      &  FID         & SOA-C & SOA-I   \\	
			\hline
			
			$\eta_1=0$ & 4.88$\pm$ 0.07 &    $16.20$  & $40.28$  &  $58.84$      \\		
			$\eta_1=0.1$  & 4.93$\pm$ 0.10 &    $15.08$  & $42.21$  &  $62.84$      \\ 		
			$\eta_1=0.5$ & 4.92 $\pm$ 0.09  &     $15.44$   & $41.00$  &  $62.12$     \\				
			$\eta_1=1.0$  & 4.96 $\pm$ 0.04  &     $15.14$   & $42.47$  &  $62.20$    \\ 			
			$\eta_1=5.0$  & 4.66 $\pm$ 0.02 &    $19.45$  &  $35.22$   & $50.63$    \\ 				
			
			\bottomrule[1.15pt]	
			
	\end{tabular} }
	\label{ablation_parameter_1} 
\end{table}

\begin{table}[]
	\centering
	\caption{IS $\uparrow$  and FID $\downarrow$    of  different values of        $\eta_2$  on   CUB-Bird dataset.    SOA-C  $\uparrow$  and SOA-I $\uparrow$  of  different values of       $\eta_2$   on    MS-COCO dataset. }
	\setlength{\tabcolsep}{3mm}{
		\begin{tabular}{c|c c | c c}
			\toprule[1.15pt]
			\multirow{2}{*}{Method}   &
			
			\multicolumn{2}{|c|}{CUB-Bird} &\multicolumn{2}{|c}{MS-COCO}  \\
			\cline{2-5} 	
			& IS      &  FID         & SOA-C & SOA-I   \\	
			\hline
			
			$\eta_2=0$ & 4.84$\pm$ 0.05 &    $16.08$  & $40.28$  &  $57.29$      \\			
			$\eta_2=0.1$  & 4.90$\pm$ 0.09 &    $16.05$  & $43.70$  &  $59.01$      \\ 		
			$\eta_2=0.5$ & 4.94 $\pm$ 0.04  &     $15.62$   & $41.82$  &  $61.94$     \\				
			$\eta_2=1.0$  &4.96 $\pm$ 0.04  &     $15.14$   & $42.47$  &  $62.20$    \\ 			
			$\eta_2=5.0$  & 4.71 $\pm$ 0.10 &    $18.33$  &  $36.62$   & $52.67$    \\ 				
			
			\bottomrule[1.15pt]	
			
	\end{tabular} }
	\label{ablation_parameter_2} 
\end{table}

\textbf{Hyperparameter \textbf{$\gamma$}.}  
	The  threshold value $\gamma$   seperates the easy   and hard parts in ALR loss.  
	We set   $\gamma$   to different values and observe the performance of    ALR-GAN, as  reported  in  Table~\ref{ablation_parameter}, from which we see that ALG-GAN achieves the  best  performance  when $\gamma=0.2$.  
	When $\gamma=0$  or $\gamma=1.0$,   there is no weight balance mechanism for the  simple   and hard parts, which results  in  a moderate drop in performance.  When $\gamma=0$  or $\gamma=1.0$, the attention weight  in   the ALR loss is still obtained through model learning. As shown in Table~\ref{ablation_parameter},   ALG-GAN still maintains appropriate performance.  

\textbf{Hyperparameters  \textbf{$\eta_1$} and \textbf{$\eta_2$}.} The  LVR loss  includes  the  Perception  Refinement (PR)  loss   and     Style  Refinement (SR) loss.   Hyperperameters  \textbf{$\eta_1$}  and  \textbf{$\eta_2$} are used to adjust the training weights of the PR   and SR loss, respectively. 
	The numerical experimental results of parameter sensitivity  are  reported  in Tables~\ref{ablation_parameter_1} and  \ref{ablation_parameter_2}. (i) When $\eta_1=0$ or $\eta_2=0$, the performance of the model degrades;  it means that  two sub-loss functions in LVR loss can  help   improve the quality of the synthesized image; (ii)  When $\eta_1 \in [0.1, 1.0]$ or $\eta_2 \in [0.1, 1.0]$, the performance of the model is stable. The values of the two parameters in  $[0.1, 1.0]$  will not cause dramatic changes in model performance; (iii)  When   the two parameters are too large, the performance  is  significantly degraded. This is because the goal of the GAN  is to learn the real image distribution and generate diverse images. This is a kind of global distribution learning based on local sampling.  Although such consistency loss as LVR loss   helps to improve the quality of the generated image, if the training weight is too large, the model will fall into the strong constraint of local samples. This destroys the learning mechanism of GANs and is prone to mode collapse.

\textbf{Hyperparameter   \textbf{$m$}.}   We discuss the effect of the number of generators on the quality of the synthesized  image.   The results are  reported  in Table~\ref{ablation_parameter1}.  The values $m=0, 1, 2$, respectively, indicate  $1$, $2$, and $3$ generators.  We can clearly see that the quality of the generated images increases with the number of generators. Due to memory limitations and fair comparison with other SOTA methods,   ALR-GAN contains three generators, i.e., $m=2$.

\textbf{Hyperparameter   \textbf{$\lambda_1$}.}   We  further   show    the   sensitivity of   $\lambda_1$, as shown  in Table~\ref{ablation_parameter_3}.   In     ALR-GAN,    $\lambda_1$ is  used to adjust the training weight of   the image  reconstruction loss $L^{REC}_{i}$ (Eq.~\ref{eq8}) in   $L_G$ (Eq.~\ref{eq121}).  The goal of the loss function $L^{REC}_{i}$  is to provide high-quality real image features to the ALR mechanism. When $\lambda_1=0$,   the loss function $L^{REC}_{i}$ is removed,  at which point the image encoder can easily lose important visual or layout information. In this way, low-quality real image features will mislead the generator to depict poor-quality layout structure and visual semantics.  When $\lambda_1 =0.001~or~0.1$,   the performance of the model is stable.  With the increase of   $\lambda_1$, the performance of the model decreases. This   is   similar to   model performance when    $\eta_1$ and $\eta_2$ are increased. If the training weight is too large, the model will fall into the strong constraint of local samples.  The training center of   ALR-GAN is also transferred to the learning of sample consistency, which weakens the GAN's learning of distribution, destroying its learning mechanism, and making it prone to mode collapse.

\begin{table}[]
	\centering
	\caption{IS $\uparrow$  and FID $\downarrow$    of  different values of     hyperparameter  $\lambda_1$  on   CUB-Bird dataset.    SOA-C  $\uparrow$  and SOA-I $\uparrow$  of  different values of    hyperparameter  $\lambda_1$   on    MS-COCO dataset. }
	\setlength{\tabcolsep}{3mm}{
		\begin{tabular}{c|c c | c c}
			\toprule[1.15pt]
			\multirow{2}{*}{Method}   &
			
			\multicolumn{2}{|c|}{CUB-Bird} &\multicolumn{2}{|c}{MS-COCO}  \\
			\cline{2-5} 	
			& IS      &  FID         & SOA-C & SOA-I   \\	
			\hline
			
			$\lambda_1=0$      &  4.44 $\pm$ 0.10     &      $23.17$       &   $28.66$    &  $45.10$       \\			
			$\lambda_1=0.01$   &  4.90$\pm$ 0.05      &      $15.12$       &   $41.38$    &  $61.64$      \\ 					
			$\lambda_1=0.1$    &  4.96 $\pm$ 0.04     &       $15.14$      &   $42.47$    &  $62.20$    \\ 
			$\lambda_1=1.0$    &  4.81 $\pm$ 0.10     &       $16.33$      &   $39.03$    &  $56.37$     \\				
			$\lambda_1=10.0$   &  4.55 $\pm$ 0.02     &      $21.73$       &   $35.10$   & $48.79$    \\ 				
			
			\bottomrule[1.15pt]	
			
	\end{tabular} }
	\label{ablation_parameter_3} 
\end{table}

\section{Conclusions}
In this paper, we presented a Text-to-Image generation model,  ALR-GAN, to improve the layout of    synthesized images.   
ALR-GAN includes an  ALR module   and     LVR loss.
The ALR module combined the   proposed ALR loss  adaptively refined  the layout structure  of the synthesized image. 
Based on the refined layout, the LVR  loss   further refined  the visual representation within the  layout area.
Experimental results and analysis demonstrated the effectiveness of these proposed  schemes, and the  ALR  module and   LVR loss     enhanced the performance of  other  GAN-based T2I methods.

\ifCLASSOPTIONcaptionsoff
  \newpage
\fi

{\small
	\bibliographystyle{ieee_fullname}
	\bibliography{egbib}
}

\end{document}